\documentclass[jimaging,article,accept,pdftex,moreauthors]{Definitions/mdpi}
\usepackage{algpseudocode}
\usepackage{amsmath}
\usepackage{amssymb}
\usepackage{graphicx}
\usepackage{balance}
\usepackage{comment}
\firstpage{1}
\pubvolume{8}
\issuenum{7}
\articlenumber{193}
\pubyear{2022}
\copyrightyear{2022}
\externaleditor{Academic Editors: Raimondo Schettini, Cosimo Distante and Ioannis Pratikakis}
\datereceived{4 June 2022}
\dateaccepted{7 July 2022}
\datepublished{11 July 2022}
\hreflink{https://doi.org/10.3390/\linebreak jimaging8070193} 

\Title{Dynamic Label Assignment for Object Detection by Combining Predicted IoUs and Anchor IoUs}

\TitleCitation{Dynamic Label Assignment for Object Detection by Combining Predicted IoUs and Anchor IoUs}

\Author{Tianxiao Zhang 
$^{1}$, Bo Luo $^{1}$, Ajay Sharda $^{2}$ and Guanghui Wang $^{3,}$*}

\AuthorNames{Tianxiao Zhang, Bo Luo, Ajay Sharda and Guanghui Wang}

\AuthorCitation{Zhang, T.; Luo, B.; Sharda, A.; Wang, G.}

\address{%
$^{1}$ \quad Department of Electrical Engineering and Computer Science, University of Kansas, Lawrence, KS 66045, USA; tianxiao@ku.edu (T.Z.); bluo@ku.edu (B.L.)\\
$^{2}$ \quad Department of Biological and Agricultural Engineering, Kansas State University, Manhattan, KS 66506, USA; asharda@k-state.edu\\
$^{3}$ \quad Department of Computer Science, Toronto Metropolitan University, Toronto, ON M5B 2K3, Canada}

\corres{Correspondence: wangcs@ryerson.ca}

\abstract{Label assignment plays a significant role in modern object detection models. Detection models may yield totally different performances with different label assignment strategies. For anchor-based detection models, the IoU (Intersection over Union) threshold between the anchors and their corresponding ground truth bounding boxes is the key element since the positive samples and negative samples are divided by the IoU threshold. Early object detectors simply utilize the fixed threshold for all training samples, while recent detection algorithms focus on adaptive thresholds based on the distribution of the IoUs to the ground truth boxes. In this paper, we introduce a simple while effective approach to perform label assignment dynamically based on the training status with predictions. By introducing the predictions in label assignment, more high-quality samples with higher IoUs to the ground truth objects are selected as the positive samples, which could reduce the discrepancy between the classification scores and the IoU scores, and generate more high-quality boundary boxes. Our approach shows improvements in the performance of the detection models with the adaptive label assignment algorithm and lower bounding box losses for those positive samples, indicating more samples with higher-quality predicted boxes are selected as positives.}


\begin{document}

\section{Introduction}

Object detection is a fundamental problem in computer vision that simultaneously classifies and localizes all objects in images or videos \cite{nguyen2021pulmonary,zhang2019real,dewi2020taiwan}. With the fast development of deep learning, object detection has achieved great success and been applied to many real-world tasks such as object tracking \cite{bharati2018real,zhang2020efficient}, image classification \cite{cen2021deep,patel2022discriminative,ma2022semantic}, segmentation \cite{he2017mask,he2021sosd}, self-driving \cite{hemmati2022adaptive}, and medical image analysis \cite{li2021colonoscopy,gosavi2022label}. Generally speaking, the detection models could be categorized as two-stage detectors and one-stage detectors. The two-stage detectors utilize Region Proposal Networks (RPN) \cite{ren2015faster} to select the anchors with high probability containing some objects and refine them. Those refined anchors are employed for classification and regression in the second stage. While one-stage detection models directly classify and regress the anchors to output the final results. The two-stage detection models often have higher accuracy while lower speed for inference compared to the one-stage detectors. RetinaNet \cite{lin2017focal} discovers that the class imbalance of positives and negatives is the reason for the accuracy gap between two-stage models and one-stage models and proposes focal loss \cite{lin2017focal} to solve this problem and close the accuracy gap between two types of models while still maintaining fast inference time of one-stage detectors. Recently, one-stage detection paradigms have a dominating influence for their high accuracy and low latency. Similar to image segmentation, object detection requires fine-grained details to accurately localize the bounding box of the object. Thus, recent object detectors frequently exploit Feature Pyramid Networks (FPN) \cite{lin2017feature} so that large feature maps with fine details could recognize the small objects, while small feature maps with large receptive field could detect the large objects.

Label assignment is to divide the samples into positives and negatives, which is essential to the success of object detection models. For anchor-based models, the core element for label assignment is the threshold for the division of positive samples and negative samples. After we calculate the Intersection over Union (IoU) between anchors and ground truth bounding boxes, the positive samples are those anchors whose IoUs are larger than the threshold, while others are negatives or ignored. Sometimes the detection models design two thresholds, one for positive and the other for negative. The anchors whose IoUs are larger than the positive threshold are positive samples, while the anchors whose IoUs are smaller than the negative threshold are negative samples. Those anchors whose IoUs are between the positive threshold and negative threshold are ignored during the training process. Those early detection models \cite{ren2015faster,liu2016ssd} utilize fixed threshold to divide the positives and negatives. However, the algorithms with the fixed threshold for dividing the positives and negatives ignore the differences between objects for their various shapes and sizes. For instance, Some large or square-shaped objects frequently have more high-quality anchors corresponding to them while some small and slender objects which have an extreme ratio of width and height often have low-quality anchors matched to them. Simply using a fixed IoU threshold for label assignment might deteriorate the performance of detecting the objects without regular shapes. Although we could design more anchors with various shapes and sizes to alleviate the problem, the overhead and computational cost are also increased, which is not desired for the requirements that prefer low latency and less computational cost.

\begin{figure}[H]
\includegraphics[width=0.9\linewidth]{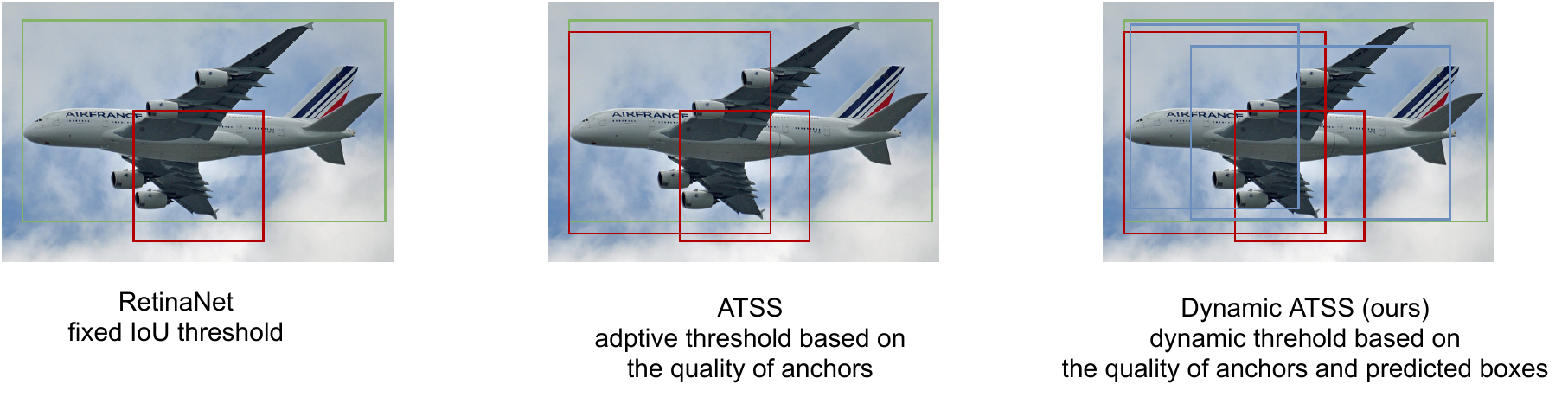}\hfill
\caption{Illustration of label assignment strategies of RetinaNet, ATSS, and our method. The green box is the ground truth bounding box and the red and blue boxes are anchors and predicted boxes by corresponding anchors, respectively. Left: RetinaNet employs a fixed IoU threshold of 0.5 to select the positive anchors. If the IoU between the anchor (red box) and the ground truth (green box) is larger than 0.5, the anchor is denoted as positive to the ground truth. Middle: ATSS calculates the adaptive IoU thresholds based on the distribution of the anchors for each ground truth. If the ground truth has the most high-quality anchors (IoU to the ground truth is high), the IoU threshold for the ground truth is also high. Otherwise, the IoU threshold is low. Right: Our method also considers the predicted box (blue box) for each anchor (red box). Even though some anchors do not satisfy the IoU threshold, they might meet the threshold considering the predicted boxes and the predicted boxes could assist the model to select the positive anchors more accurately.}
\label{fig:0}
\end{figure}

Recently, more and more adaptive label assignment strategies (e.g., ATSS \cite{zhang2020bridging}) have been proposed to adaptively calculate the threshold. These algorithms adaptively select positive samples and negative samples based on the IoU distribution between anchors and ground truth bounding boxes so that the ground truth bounding boxes that have more high-quality anchors corresponding to them will have a higher IoU threshold and those which have the most low-quality anchors corresponding to them will have a low IoU threshold, as illustrated in Figure \ref{fig:0}. Nevertheless, adaptive assignment methods do not assign positives and negatives based on the predictions which are more accurate to represent the training status. Due to the discrepancy between the classification and localization, classification scores cannot precisely correspond to the localization quality, while NMS (non-maximum suppression) supposes classification scores represent the localization quality and filters duplicates so that only the samples with high classification scores will be kept. However, if classification scores cannot accurately represent the localization quality, the high-quality bounding boxes might be eliminated and some low-quality bounding boxes might be kept. However, fixed anchors cannot guarantee the quality of the predicted bounding boxes. 

Therefore, introducing predictions to instruct label assignment is an effective approach to include the anchors which could generate high-quality predictions as the positives. For some models using dynamic label assignment strategies, predictions are utilized to guide the label assignment during training since they represent the true training status. Nonetheless, the predictions in the early stage of training are inaccurate for both classification and bounding box regression; thus, they are inappropriate to instruct the label assignment if we directly apply them to dividing positive samples and negative samples. Adding distances to the ground truth centers as prior is proposed in the algorithm which utilizes predictions to weight the positive samples. Those prediction-based label assignment strategies exploit the distances or bounding boxes as the prior so that the positives are limited to the ground truth bounding boxes or those positives samples that are closer to the centers of the ground truth bounding boxes have more weights during the training process \cite{zhu2020autoassign,ge2021lla}. The predictions (classification scores or predicted boxes) and the distances are two different ``domains'', so they could not be naturally combined. Thus, Autoassign \cite{zhu2020autoassign} designed a center weighting module to solve this problem; however, the module may be sub-optimal due to the assumption that the samples closer to centers of the ground truth would have more weights. MAL \cite{ke2020multiple} employs ``All-to-Top-1'' strategy which includes enough anchors to learn the detector in the early training stage, and the number of anchors gradually decrease with the training process going on and finally only one optimal anchor is utilized. However, ''All-to-Top-1'' \cite{ke2020multiple} reduces the number of anchors in the bag based on iterations instead of predictions. Thus, the training may not be optimal since the number of anchors in the bag is not controlled by predictions and might not satisfy the training status.

In this paper, we propose a simple and effective method that directly combines the predicted IoUs between the predicted bounding boxes and the ground truth bounding boxes, and the anchor IoUs between the anchors and the ground truth bounding boxes. According to the adaptive models, the adaptive threshold is attained according to the statistical properties of the IoUs between the candidate anchors and the ground truth bounding boxes. Our method computes the distribution of the predicted IoUs and the anchor IoUs separately and then attains the combined parameters by simply adding them, respectively. Finally, the combined threshold is computed by the combined distribution parameters. Since the predictions are involved in the label assignment, soft targets (predicted IoUs between the predicted bounding boxes and the ground truth boxes) are more appropriate than the hard target (label 1) for positives in classification loss. QFL \cite{li2020generalized} and VFL \cite{zhang2021varifocalnet} are commonly utilized classification loss with soft target. Both of them could further boost the performance of our proposed method. In addition, we replace the Centerness branch with the IoU branch for better accuracy. 


In our work, we demonstrate the importance of using predictions to include samples with higher-quality predicted boxes as positives for label assignment and propose a simple and effective approach to introduce the predictions to the definition of positive and negatives. Our method could naturally introduce the predictions to label assignment algorithms and the anchors could perform as prior for predictions. The experiments on COCO dataset~\cite{lin2014microsoft} illustrate the effectiveness of our method without extra cost. Designing object detection models dynamically based on training status are a trend recently. While directly using the predictions is unreasonable due to the inaccurate predictions in the early stage of training. Thus, how to create prior to restrict the predictions is important but under-explored. Our idea of exploiting anchor IoUs as prior to combine with predicted IoUs could be a good choice and might inspire more related works.


\section{Related Works}

\subsection{Object Detection}
Object detection could be categorized as two-stage approaches and one-stage approaches \cite{li2020object,ma2020mdfn,xu2020adaptively,mo2021stereo}. Two-stage detection models \cite{ren2015faster,cai2016unified,dai2016r,lin2017feature} first employ Region Pyramid Networks (RPN) \cite{ren2015faster} to select anchors with high possibility to have high IoUs with some ground truth objects and refine those candidate anchors. Then the refined anchors are fed into the second stage to be classified and further regressed. While one-stage detectors~\cite{liu2016ssd,redmon2016you,redmon2017yolo9000,redmon2018yolov3,lin2017focal} directly classify and regress the anchors without selecting and refining some candidates. Two-stage detectors often yield higher accuracy but lower speed compared to one-stage counterparts. With the emergence of RetinaNet \cite{lin2017focal}, the accuracy discrepancy between the one-stage and two-stage models has been reduced by introducing focal loss to suppress the loss of easy samples so that one-stage methods could achieve both high accuracy and low latency. Thus, one-stage approaches are dominating current object detection models.

With the appearance of anchor-free models \cite{law2018cornernet,duan2019centernet,zhou2019bottom,zhu2019feature,tian2019fcos}, the pre-defined anchors are no longer required for a well-performed detection model. The anchor-free models either regress the bounding box from anchor points (feature points) \cite{zhu2019feature,tian2019fcos} or predicts some special points of the ground truth object such as corners or extreme points of the boundary boxes of the objects \cite{law2018cornernet,duan2019centernet,zhou2019bottom} and finally constructs the predicted bounding boxes from those special points. Recently, some object detection models \cite{patel2022aggregating,carion2020end,zheng2020end,zhu2020deformable,ma2021miti} are enhanced by attention modules with Transformers \cite{vaswani2017attention}, which are originally invented for natural language processing. DETR \cite{carion2020end} first introduces the Transformers to the head of detection models, which is also anchor free. Nonetheless, due to the global attentions utilized in Transformers and the large resolution of the images for object detection, DETR requires much longer epochs than its CNN counterparts to convergence. Thus, recent algorithms \cite{meng2021conditional,gao2021fast} attempt to design fast training convergence DETR to speed up the training process.

\subsection{Label Assignment}

The label assignment is the core factor for the performance of the detection models and how to divide positive samples and negative samples would determine how the networks learn and converge. For RPN training in Faster R-CNN \cite{ren2015faster}, the pre-defined anchors whose IoUs with any ground truth bounding box higher than 0.7 will be defined as positives, and those whose IoUs with all ground truth objects lower than 0.3 are negatives. The anchors with IoUs between 0.3 and 0.7 are ignored during training and the networks do not learn from those anchors. For SSD \cite{liu2016ssd}, the threshold of IoU between the positive anchor boxes and the ground truth objects is 0.5. RetinaNet \cite{lin2017focal} assigns an anchor box to a ground truth object if the IoU between them is higher than or equal to 0.5, and assigns an anchor box as a negative sample if the IoUs between the anchor and all ground truth boxes are lower than 0.4. Others are ignored during training. The aforementioned strategies are traditional label assignment approaches with fixed thresholds for dividing positives and negatives. Even though those detection models with fixed thresholds are still effective for label assignment, they ignore the differences between various object samples for their shapes, sizes and number of corresponding positive anchors. For instance, with fixed thresholds for label assignment, the objects with square-like shapes or large sizes might have more high-quality anchors corresponding to them so that they have more positive anchors during training, while some objects with slender shapes or small sizes might have the majority of low-quality corresponding anchor boxes; thus, they correspond to fewer positive anchors during training. So during training, the networks will be biased toward the objects with balanced ratios of width and height or large sizes and the performance of some slender or small objects will be compromised.

Recently, researchers focus on designing adaptive thresholds and gradually discard the fixed thresholds for label assignment. ATSS \cite{zhang2020bridging} computes the adaptive thresholds by calculating the mean and standard deviation according to the distribution of the IoUs between the candidate anchors and the ground truth objects. PAA \cite{kim2020probabilistic} attains the anchor scores by combining the classification scores and localization scores, and then the selected anchor candidates which are based on top anchor scores are fitted into Gaussian Mixture Model (GMM) and the GMM is optimized by Expectation-Maximization algorithm probabilistically. The candidate anchor boxes are separated by the boundary schemes as positives and negatives.

Using predictions to guide the label assignment could be more accurate since the pre-defined anchors might not accurately reflect the actual training status. Nevertheless, predictions in the early training stage are inaccurate and unreasonable to instruct the label assignment. FreeAnchor \cite{zhang2019freeanchor} exploits maximum likelihood estimation (MLE) to model the training process so that each ground truth could have at least one corresponding anchor with both high classification score and localization score. To solve the inaccurate predictions in the early epoch of training, the mean-max function is proposed so that almost all candidate anchors could be used in training the network and the best anchor could be selected from candidates when the training is sufficient, which is similar to ``All-to-Top-1'' mechanism in MAL \cite{ke2020multiple}. MAL \cite{ke2020multiple} employs predictions from classification and localization as the joint confidence for evaluation of the anchors. To alleviate the sub-optimal anchor-selection problem, MAL perturbs the features of selected anchors based on the joint confidence to suppress the confidence of those anchor candidates so that other anchors could have a chance to be selected and participate in the training process. In addition, MAL proposes ``All-to-Top-1'' strategy for anchor selection that includes enough anchors to be involved in training and gradually decreases the number of anchors in the bag to 1 with the training process going on. However, linearly decreasing the number of anchors in the bag is irrelevant to the prediction status and perturbing the features of selected anchors introduces randomness which is also not correlated to the current predictions. Thus, this strategy might not be optimal to train the network.

Autoassign \cite{zhu2020autoassign} introduces Center Weighting as the prior to address the unreasonable predictions in the early training phase, which indicates that the samples closer to the centers of ground truth would have more weights. Nonetheless, the prediction confidence and the distance to the ground truth centers are two different ``domains'' and the combination might be sub-optimal even though Autoassign models the distances with Gaussian-shape weighting function \cite{zhu2020autoassign}. Our model naturally combines the predicted IoUs (IoUs between the predicted boxes and the ground truth boxes) and the anchor IoUs (IoUs between the anchor boxes and the ground truth boxes) as the combined IoUs. Since they are both IoUs to the ground truth boxes, they could be naturally combined together without any complex functions or weights. In the early training stage, the anchor IoUs dominate the training process due to the random initialization. As the training process proceeds, the predicted IoUs will gradually dominate the combined IoUs and more samples with high-quality bounding boxes will be selected as positives.

\section{Proposed Approach}

The adaptive label assignment strategies frequently divide the positive and negative samples by calculating the statistical parameters (e.g., mean and standard deviation) based on the candidate anchors or anchor bags which are selected according to the Euclidean distances between the centers of the anchors to the centers of the ground truth bounding boxes. After the candidate anchors are selected based on their positions to the ground truth boxes, the adaptive thresholds are computed based on the distribution of their IoUs to the corresponding ground truth bounding boxes. In the paper, ATSS \cite{zhang2020bridging} is utilized as an example to illustrate the adaptive label assignment method and our proposed approach.

\subsection{Revisit ATSS}

ATSS (Adaptive Training Sample Selection) \cite{zhang2020bridging} makes an empirical analysis of the anchor-free approaches and anchor-based approaches and concludes that how to divide the positive samples and negative is the significant difference between anchor-based models and anchor-free models. Thus, ATSS proposes an algorithm that calculates the adaptive thresholds for defining the positives and negatives based on the candidate anchors which are selected according to the Euclidean distances between the centers of the anchors to the centers of the ground truth bounding boxes. The ATSS algorithm is shortly summarized as~below:

For each GT (ground truth) box, 

\begin{enumerate}
\item[1:] Compute the Euclidean distances between the centers of all anchors and the center of the GT.
\item[2:] Select $k$ anchors with the smallest distances for each feature pyramid level; if the number of feature levels is $L$, the number of total candidate anchors corresponding to the GT is $kL$.
\item[3:] Compute the IoUs between $kL$ candidate anchors and the GT box. Then calculate the mean and standard deviation of those IoUs.
\item[4:] The adaptive threshold is mean+std. If one anchor has IoU with the GT larger than or equal to mean+std, it is candidate positive of the GT, else it is negative.
\item[5:] Only the candidate positives whose centers are inside the GT box would be the final positives of the GT, others are negatives.
\end{enumerate}

ATSS computes the threshold adaptively according to the shapes and sizes of the GT bounding boxes. If the GT boxes are large or square-like, the threshold will be higher since there are more high-quality anchors corresponding to them. If the GT boxes have slender shapes or small sizes, the threshold will be lower due to most low-quality anchors corresponding to them. Nevertheless, ATSS only computes adaptive thresholds according to the relationship between anchors and GT boxes. It merely relies on the anchors and ignores the actually predicted bounding boxes. In other words, the anchor with the highest IoU to the GT box cannot guarantee its predicted bounding box also has the highest IoU to the GT among all positive anchors. Thus, some samples with high-quality predicted bounding boxes might be defined as negative samples whose classification target is 0. Thus, the performance of high-quality bounding boxes is affected. We will demonstrate it in the experiments. Using predicted information may improve the accuracy of defining the positives and negatives since predictions can represent the real training status of each sample. However, directly using the predictions might not be appropriate since the predictions in the early training stage are unreasonable to instruct the positive and negative definitions. Thus, we propose a simple and effective approach to address this problem by combining the predicted IoUs to the GT and the pre-defined anchor IoUs to the GT for each training sample.

\subsection{Dynamic ATSS}

\begin{figure}[H]
\includegraphics[width=0.9\linewidth]{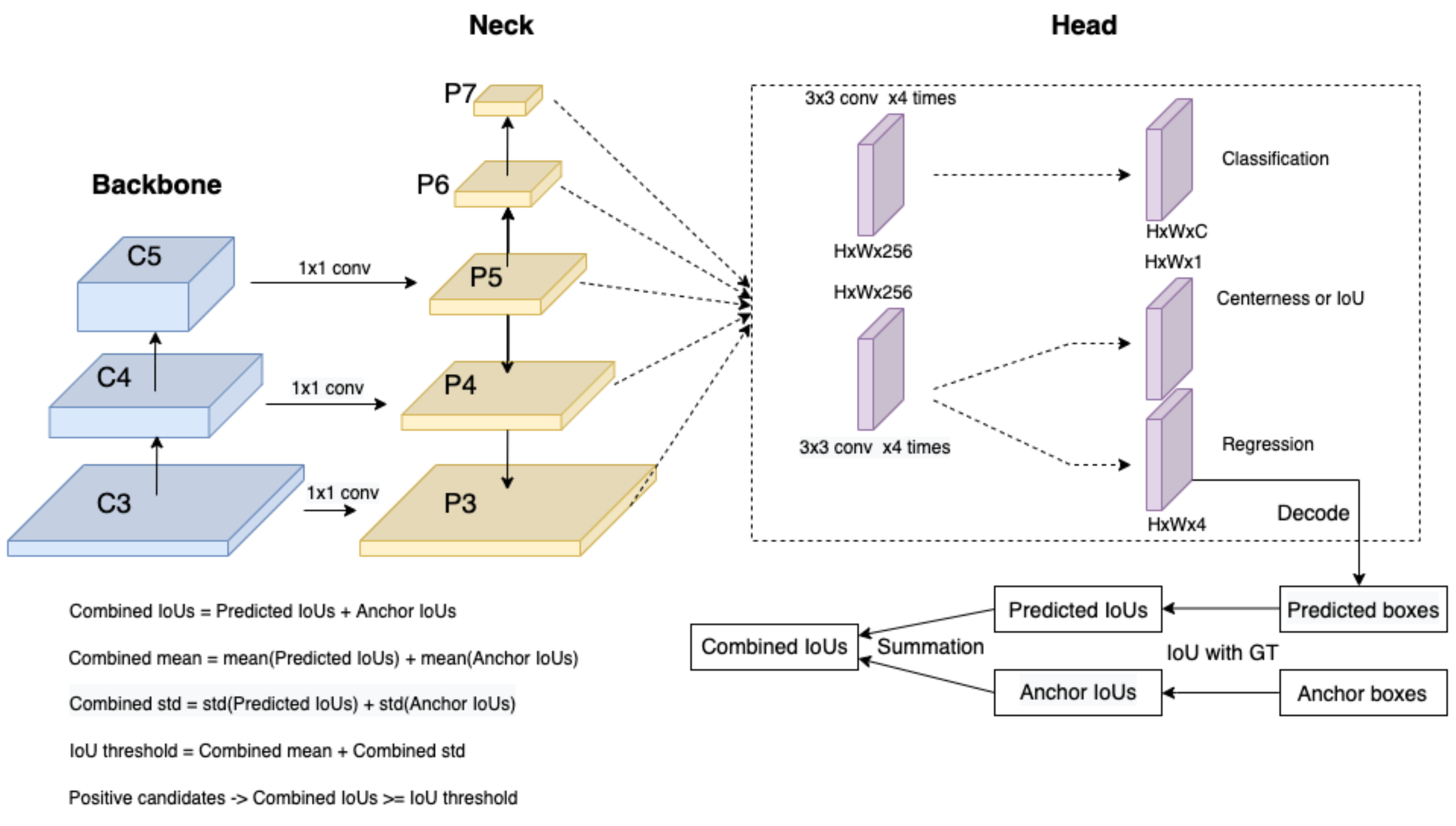}\hfill
\caption{The 
network structure of our model. The structure of our model is the same as ATSS~\cite{zhang2020bridging}, which includes a CNN backbone \cite{he2016deep}, an FPN neck \cite{lin2017feature} and a head that has two branches for classification and regression, respectively. Our approach would employ the predicted boxes which are decoded from the regression branch. Then the predicted IoUs and the anchor IoUs are attained by calculating the IoUs between the predicted boxes and the GTs, and the IoUs between the anchor boxes and the GTs, respectively. Finally, the Combined IoUs (CIoUs) are computed by summing the predicted IoUs and the anchor IoUs. The same calculation is implemented to attain the combined mean and combined std. The IoU threshold is computed by the summation of combined mean and combined std, and the positive candidates are defined as the samples whose Combined IoUs are larger than or equal to the IoU threshold. The positive candidates are restricted inside the ground truth bounding boxes as the final positive samples.}
\label{fig:1}
\end{figure}

We propose Dynamic ATSS, which introduces the predictions to the anchors for label assignment. In the early training phase, the predictions are inaccurate due to random initialization. Thus, the anchors would act as prior to instruct the label definition. The predictions gradually dominate the combined IoUs and would lead the label assignment with the training proceeding and the predictions being improved. The network structure is illustrated in Figure \ref{fig:1}. The network structure is the same as ATSS, which has a CNN backbone \cite{he2016deep}, an FPN neck \cite{lin2017feature} and a shared head which has two branches for classification and regression, respectively, while our proposed approach would extract the regression results and decode the regression offsets to the coordinates of bounding boxes, and finally calculate the IoUs between the decoded bounding boxes and the GTs. The predicted IoUs will be combined with anchor IoUs for selecting the positive samples.
\begin{equation} \label{eq:1}
CIoUs = PIoUs + AIoUs
\end{equation}
\begin{equation} \label{eq:2}
mean(CIoUs) = mean(PIoUs) + mean(AIoUs)
\end{equation}
\begin{equation} \label{eq:3}
std(CIoUs) = std(PIoUs) + std(AIoUs)
\end{equation}
\begin{equation} \label{eq:4}
threshold(CIoUs) = mean(CIoUs) + std(CIoUs)
\end{equation}

Equations (\ref{eq:1})--(\ref{eq:4}) illustrate our proposed methods. $PIoUs$ is the predicted IoUs between the predicted bounding boxes and the ground truth boxes, and $AIoUs$ indicates the anchor IoUs between the pre-defined anchor boxes and the ground truth boxes. $CIoUs$ indicates the combined IoUs which is the summation of predicted IoUs and anchor IoUs. When calculating the mean and standard deviation for $CIoUs$, we compute them for $PIoUs$ and $AIoUs$ separately and sum them together, as illustrated in Equations (\ref{eq:2}) and (\ref{eq:3}). Finally, the thresholds of $CIoUs$ for defining the positives and negatives are computed by summation of the mean and standard deviation of $CIoUs$. Since $PIoUs$ and $AIoUs$ are both IoUs to the ground truth bounding boxes, they can be naturally combined together by summation without designing any sophisticated formula \cite{zhu2020autoassign} or reducing the number of positives during the training process \cite{ke2020multiple}. We also implement some experiments that down-weight the $AIoUs$ or up-weight the $PIoUs$ with the training proceeding. However, the simple addition of $PIoUs$ and $CIoUs$ could yield the best results, which is demonstrated in the experiments.

Why is utilizing predictions so important to guide the label assignment? The predictions are more accurate than the pre-defined anchors for defining the positives and negatives since we select the final results and implement the NMS algorithm based on the predicted results instead of the anchor boxes. We frequently design the detection models based on the assumption that the samples whose pre-defined boxes have high IoUs with the ground truth boxes are appropriate to be selected as positives or the samples whose centers are close to the centers of the ground truth objects are good candidates for positives. Once the positive samples are selected for each image, they would not be modified during the training process since the pre-defined anchor boxes or anchor points are fixed and they would not be changed according to the training status. Nevertheless, the samples with high-quality predictions might not frequently be those samples with high-quality anchor boxes or anchor points, although they have higher probabilities to generate high-quality predictions.

If we force the samples with high-quality anchor boxes or anchor points to be the positives through the entire training process, the network would focus on learning those samples even though their predictions are not good enough and ignore those samples which could generate better-predicted results but may be assigned as negatives due to the relatively low-quality anchor boxes or anchor points. While if the predictions are introduced to assist the definition of positive samples and negative samples, we could select more samples with high-quality predictions as positives and further improve those samples. From Table \ref{table:1}, simply adding predicted IoUs to anchor IoUs could yield better results and generate higher-quality predictions. The anchor IoUs are also necessary for our approach due to the random initialization of the network and they can act as prior. In our method, the predictions and the prior are both IoUs to the ground truth bounding boxes; thus, they can be naturally combined together by addition without any special design, which is shown in Figure \ref{fig:1}.

\begin{table}[H]
\caption{The 
effectiveness of the proposed method.}
\newcolumntype{C}{>{\centering\arraybackslash}X}
\begin{tabularx}{\textwidth}{CCCCCCC}
\toprule
\textbf{Model} & \textbf{AP} & \textbf{AP50} & \textbf{AP75} & \textbf{APs} & \textbf{APm} & \textbf{APl} \\
\midrule
ATSS & 39.06 & 57.11 & 42.49 & 22.33 & 43.27 & 50.23 \\
\midrule
ATSS+CIoUs & \textbf{39.75} & 57.43 & 43.08 & 23.03 & 43.83 & 52.27 \\
\midrule
ATSS+QFL & 39.61 & 57.41 & 43.05 & 23.25 & 43.69 & 52.19 \\
\midrule
ATSS+VFL & 39.65 & 57.38 & 43.39 & 23.45 & 43.53 & 52.19 \\
\bottomrule

\end{tabularx}
\label{table:1}
\end{table}

\subsection{Soft Targets for Classification Loss}

With the emergence of focal loss \cite{lin2017focal}, most modern object detection models exploit focal loss for learning the class labels. Focal loss addresses the extreme imbalances between positive samples and negative samples during training and suppresses the majority of easy negative samples which could dominate the training loss due to the extremely large number of those easy negatives. The focal loss function for positive samples ($y=1$) and negative samples ($y=0$) is shown in Equation (\ref{eq:5}).
\begin{equation} \label{eq:5}
\text{FL} =
\begin{cases}
-\alpha(1-p)^\gamma log(p), & \text{$y = 1$} \\
-(1-\alpha)p^\gamma log(1-p), & \text{$y = 0$}
\end{cases}
\end{equation}

Due to the introduction of predictions for label assignment, using soft target (predicted IoUs to the ground truth boxes) is more appropriate to rank the high predicted IoUs on top of other low predicted IoUs, which is utilized in GFL \cite{li2020generalized} and VFNet \cite{zhang2021varifocalnet}. GFL is comprised of QFL and DFL for classification and regression, respectively. We employ QFL for classification in our model. QFL \cite{li2020generalized} for classification is illustrated in Equation (\ref{eq:6}).
\begin{equation} \label{eq:6}
\text{QFL} =
\begin{cases}
-|y-p|^\beta[ylog(p)+(1-y)log(1-p)], & \text{$y > 0$} \\
-p^\beta log(1-p), & \text{$y = 0$}
\end{cases}
\end{equation}

From the QFL equation, the cross-entropy loss is switched to the general form for positives ($y > 0$) since the soft targets are not equal to 1. In addition, the focal loss weights are also modified according to the soft targets.

Instead of down-weighting the losses when the classification predictions approach to the soft targets as used in QFL, VFNet \cite{zhang2021varifocalnet} exploits VFL \cite{zhang2021varifocalnet} that weights the positive losses with the soft targets to which those positives are assigned, which is demonstrated in Equation (\ref{eq:7}). By changing the weights to the IoU targets for positives ($y > 0$), the losses of positive samples with higher IoU targets would be also higher so that the network could focus on learning those high-quality positives.
\begin{equation} \label{eq:7}
\text{VFL} =
\begin{cases}
-y[ylog(p)+(1-y)log(1-p)], & \text{$y > 0$} \\
-\alpha p^\gamma log(1-p), & \text{$y = 0$}
\end{cases}
\end{equation}

In the experiments, we could empirically illustrate that our proposed approach surpasses the same model using QFL or VFL in Table \ref{table:1}. In addition, by combining our proposed method with QFL or VFL, the performance of the detection model could be further improved.

\section{Experiments}

{Implementation Details:
} The experiments are conducted on COCO dataset \cite{lin2014microsoft}, where \textit{train2017} is employed for training and \textit{val2017
} is for validation and ablation study. The training strategy exploits scheduler 1x (90 k iterations) for validation where the batch size is 16, and the initial learning rate is 0.01 and the learning rate is decreased by a factor of 10 at 60 k and 80 k iterations. The images are resized so that the maximum length of the longer side is 1333 and the length of the shorter side is 800. ResNet-50 \cite{he2016deep} is utilized for all ablation study.

\subsection{Ablation Study}

In the experiments, ATSS algorithm \cite{zhang2020bridging} is selected as an example to illustrate the effectiveness of our proposed approach. 
We introduce the predicted IoUs to the ATSS algorithm to demonstrate the effectiveness of our approach.

\subsubsection{The Effectiveness of Proposed Method}

To prove the effectiveness of our proposed method, we did several experiments based on the ATSS algorithm. The experimental results are shown in Table \ref{table:1}.

From Table \ref{table:1}, ATSS combined with our proposed CIoUs (Combined IoUs) surpasses the same model with soft targets (QFL and VFL) for classification loss. Our simple modification can boost the original ATSS algorithm by around 0.7 AP on COCO \textit{val2017} dataset, which demonstrates that using predictions could better guide the positive and negative definitions and the anchor boxes are also necessary for instructing the label assignment, especially in the early stage of the training process. By simply combining them together, the model could yield great accuracy improvement. We simply introduce CIoUs into ATSS here and the labeled target is still the hard target (1 for positives). In the following experiments, we will show that the performance could be further boosted with the soft target (QFL or~VFL).

\subsubsection{The Contribution of Each Element}

In this section, we implement the experiments by removing some of the components. Through this ablation study, we can easily recognize the contribution of each element. The experimental results are demonstrated in Table \ref{table:2}.

In Table \ref{table:2}, similar to the aforementioned Equations (\ref{eq:1})--(\ref{eq:4}), AIoUs represents the IoUs between the pre-defined anchor boxes and the ground truth bounding boxes. If only AIoUs are selected, the original ATSS is implemented. PIoUs indicate the IoUs between the predicted bounding boxes and the ground truth boxes. If both AIoUs and PIoUs are selected, our proposed Combined IoUs are implemented by summing the computed AIoUs and PIoUs. We can obviously notice that only employing PIoUs for label assignment significantly compromises the performance of the model from 39.06 AP to 29.39 AP, while simply adding the PIoUs to AIoUs for defining the positive samples and negative samples could yield around 0.7 AP improvement and the improvement happens in all metrics (AP, AP50, AP75, APs, APm, APl), which verifies that using predictions could include more candidates with high-quality predicted bounding boxes as the positive samples and finally improve the overall performance of the detection models.

\begin{table}[H]
\caption{The ablation study for each element.}
\centering
\begin{adjustwidth}{-\extralength}{0cm}
\newcolumntype{C}{>{\centering\arraybackslash}X}
\begin{tabularx}{\fulllength}{CCCCCCCCCCCC}
\toprule
\textbf{AIoUs} & \textbf{PIoUs} & \textbf{QFL} & \textbf{VFL} & \textbf{Centerness Branch} & \textbf{IoU Branch} & \textbf{AP} & \textbf{AP50} & \textbf{AP75} & \textbf{APs} & \textbf{APm} & \textbf{APl} \\
\midrule
\checkmark & & & & \checkmark & & 39.06 & 57.11 & 42.49 & 22.33 & 43.27 & 50.23 \\
\midrule
& \checkmark & & & \checkmark & & 29.39 & 46.77 & 31.13 & 21.57 & 28.38 & 37.08 \\
\midrule
\checkmark & \checkmark & & & \checkmark & & 39.75 & 57.43 & 43.08 & 23.03 & 43.83 & 52.27 \\
\midrule
\checkmark & \checkmark & \checkmark & & \checkmark & & 40.07 & 57.46 & 43.73 & 23.47 & 44.30 & 52.60 \\
\midrule
\checkmark & \checkmark & & \checkmark & \checkmark & & 39.83 & 57.45 & 43.15 & 22.75 & 44.22 & 52.88 \\
\midrule
\checkmark & \checkmark & \checkmark & & & \checkmark & \textbf{40.30 
} & 57.49 & 44.00 & 22.85 & 44.48 & 53.71 \\
\midrule
\checkmark & \checkmark & & \checkmark & & \checkmark & 40.15 & 57.37 & 43.64 & 23.51 & 44.09 & 53.21 \\
\bottomrule
\end{tabularx}
\label{table:2}
\end{adjustwidth}
\end{table}

\begin{figure}[H]
{\includegraphics[width=0.58\linewidth]{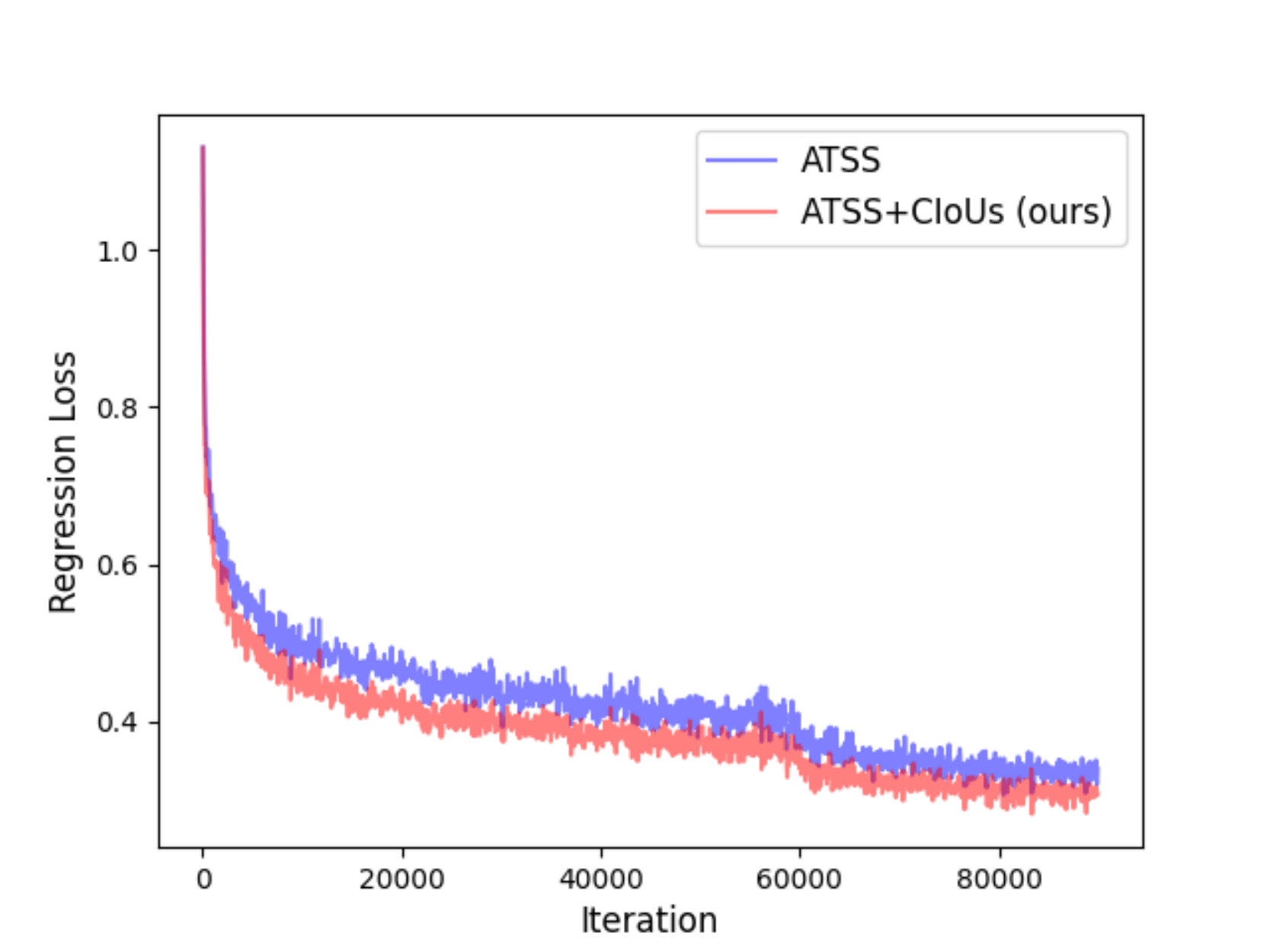}}
\caption{The regression loss of ATSS and ATSS+CIoUs.}
\label{fig:2}
\end{figure}

Figure \ref{fig:2} illustrates the regression loss of the original ATSS and ATSS with our proposed combined IoUs. From Figure \ref{fig:2}, the regression loss does not have too much difference in the early training phase for both models. While with the training process going on, our approach has lower regression loss than the original model, which indicates our model could select positive samples with higher-quality bounding boxes since more accurate predicted bounding boxes would yield lower regression loss. Additionally, the average precision for large objects (APl) is greatly improved by about 2\%.

From Table \ref{table:2}, our proposed approach (AIoUs+PIoUs) could be further improved by soft targets (QFL and VFL). The original ATSS implements Centerness as the additional branch to weight the positive samples so that the samples closer to the centers of the GTs could have relatively higher weights than those far from the centers of the GTs. After switching the Centerness to IoU (predicting IoU instead of centerness), the performance could be boosted. 

\subsubsection{Balancing Predicted IoUs and Anchor IoUs}

We also investigate how to balance the predicted IoUs and the anchor IoUs when calculating the dynamic thresholds. The experiments are illustrated in Table \ref{table:3}.

\begin{table}[H]
\centering
\caption{The ablation study on the weights for combination.}
\newcolumntype{C}{>{\centering\arraybackslash}X}
\begin{tabularx}{\textwidth}{CCCCCCCC}
\toprule
\textbf{PIoUs} & \textbf{AIoUs} & \textbf{AP} & \textbf{AP50} & \textbf{AP75} & \textbf{APs} & \textbf{APm} & \textbf{APl} \\
\midrule
1 & 1 & \textbf{39.75 
} & 57.43 & 43.08 & 23.03 & 43.83 & 52.27 \\
\midrule
0.5 & 1 & 39.43 & 57.23 & 42.92 & 23.04 & 43.56 & 51.31 \\
\midrule
1.5 & 1 & 39.55 & 57.38 & 42.63 & 23.15 & 43.28 & 51.60 \\
\midrule
1 & 0.5 & 39.42 & 57.13 & 42.46 & 22.95 & 43.04 & 51.03 \\
\midrule
1 & 1.5 & 39.51 & 57.27 & 42.68 & 22.94 & 43.31 & 51.72 \\
\midrule
D\_up & 1 & 39.22 & 56.92 & 42.50 & 22.74 & 42.98 & 51.35 \\
\midrule
1 & D\_down & 35.05 & 53.33 & 37.58 & 22.44 & 35.76 & 46.27 \\
\midrule
D\_up & D\_down & 35.64 & 53.78 & 38.09 & 22.68 & 36.57 & 48.12 \\
\bottomrule
\end{tabularx}
\label{table:3}
\end{table}

From Table \ref{table:3}, the model with the ratio between PIoUs and AIoUs being 1:1 could yield the best accuracy. D\_up indicates the dynamic weight gradually increases the weight with the training proceeding. D\_down represents the dynamic weight that gradually decreases the weight with the training proceeding. In the experiment, we utilize $iteration/max\_iter$ 
as the D\_up and $(1-iteration/max\_iter)$ as the D\_down. In the representation, $iteration$ is the current iteration and $max\_iter$ represents the total iterations the training would implement. The initial value of D\_up is close to 0 and the final value of D\_up would be close to 1 when the training is approaching to the end. While the initial value of D\_down is close to 1 and the final value of D\_down would be close to 0, which is the opposite of D\_up.

Intuitively, the weight of the predicted IoUs should gradually increase since the predicted IoUs would be more and more accurate with the training proceeding. Thus, in our experiments, we apply D\_up to the predicted IoUs and D\_down to the anchor IoUs so that the predicted IoUs would gradually take over the label assignment and the anchor IoUs would gradually fade out of the label assignment. Nonetheless, introducing the dynamic weights to PIoUs and AIoUs does not improve the performance and applying D\_down to AIoUs greatly worsen the accuracy of the detection model, which verifies that anchors still play a significant role in our method. Since PIoUs would gradually increase with the training proceeding due to increasing localization accuracy of predicted boxes, dynamic weight has no positive effect to PIoUs. In our experiments, simple summation of PIoUs and AIoUs is applied and no weights are utilized for combining them. This ablation study also demonstrates that anchor IoUs (prior) and predicted IoUs could be naturally combined together without any sophisticated formula or weights, which is simpler than using complicated formula to combine the center distance (prior) and the predictions \cite{zhu2020autoassign} or ``All-to-Top-1'' \cite{ke2020multiple} that gradually decreases the number of the anchors in the bag.

\subsection{Application to the State-of-the-Art}

The proposed approach combines the predicted IoUs with the anchor IoUs, so it could be directly applied to some state-of-the-art models that employ the ATSS algorithm for label assignment. GFLV2 \cite{li2021generalized} and VFNet \cite{zhang2021varifocalnet} are both state-of-the-art detection models that employ ATSS algorithm for defining positives, negatives, and soft targets (IoUs) for the classification loss. Instead of directly predicting the offsets of the bounding boxes, GFLV2~\cite{li2021generalized} designs a distribution-guided quality predictor to estimate the localization quality via the bounding box distributions, and VFL \cite{zhang2021varifocalnet} makes a further bounding box refinement by employing a star-shaped box feature representation. We applied our method to GFLV2 \cite{li2021generalized} and VFL \cite{zhang2021varifocalnet} and the experimental results are shown in Table \ref{table:4}.

\begin{figure}[H]
\includegraphics[width=0.58\linewidth]{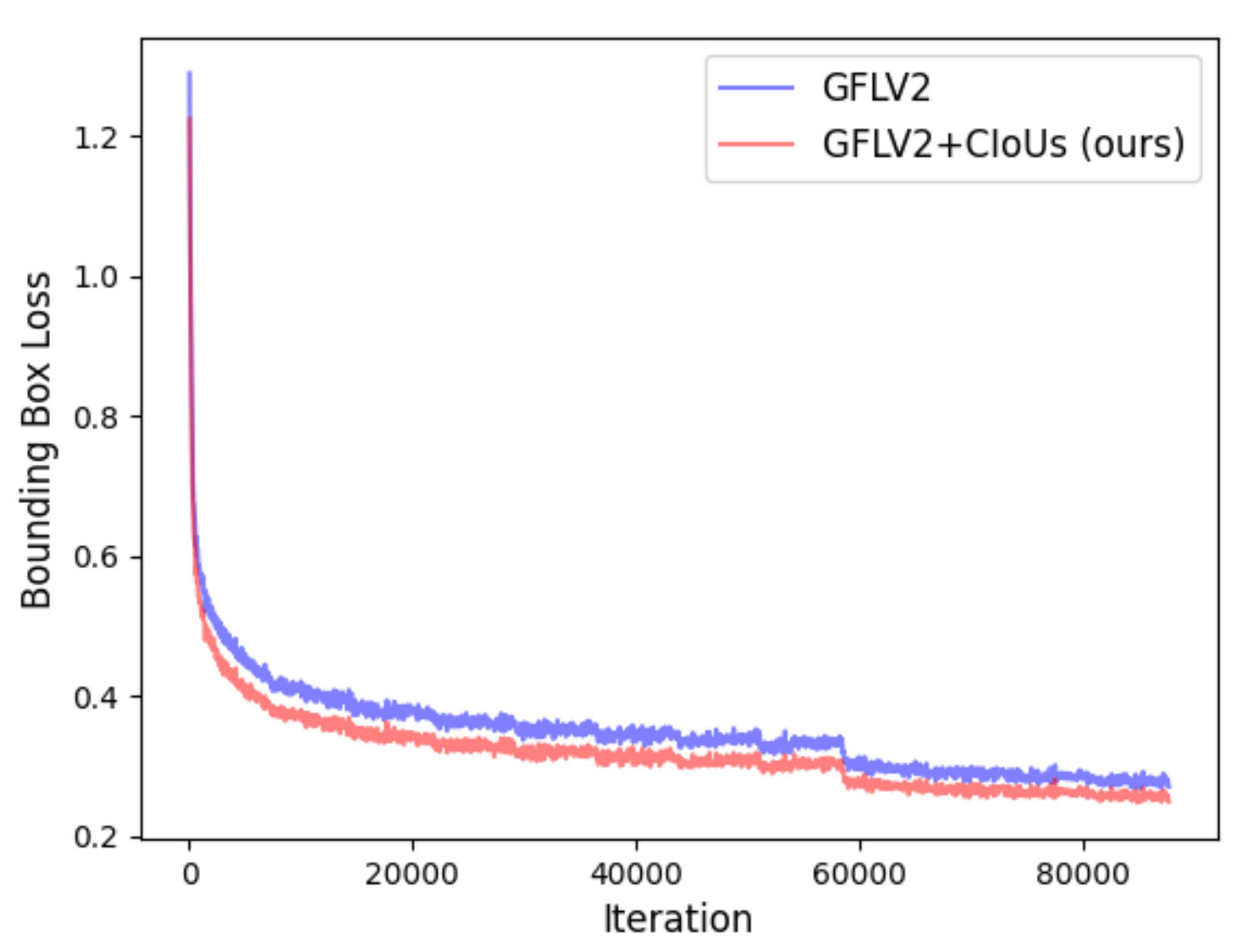}
\caption{The bounding box loss of GFLV2 and GFLV2+CIoUs.} 
\label{fig:3}
\end{figure}

\begin{figure}[H]
\includegraphics[width=0.58\linewidth]{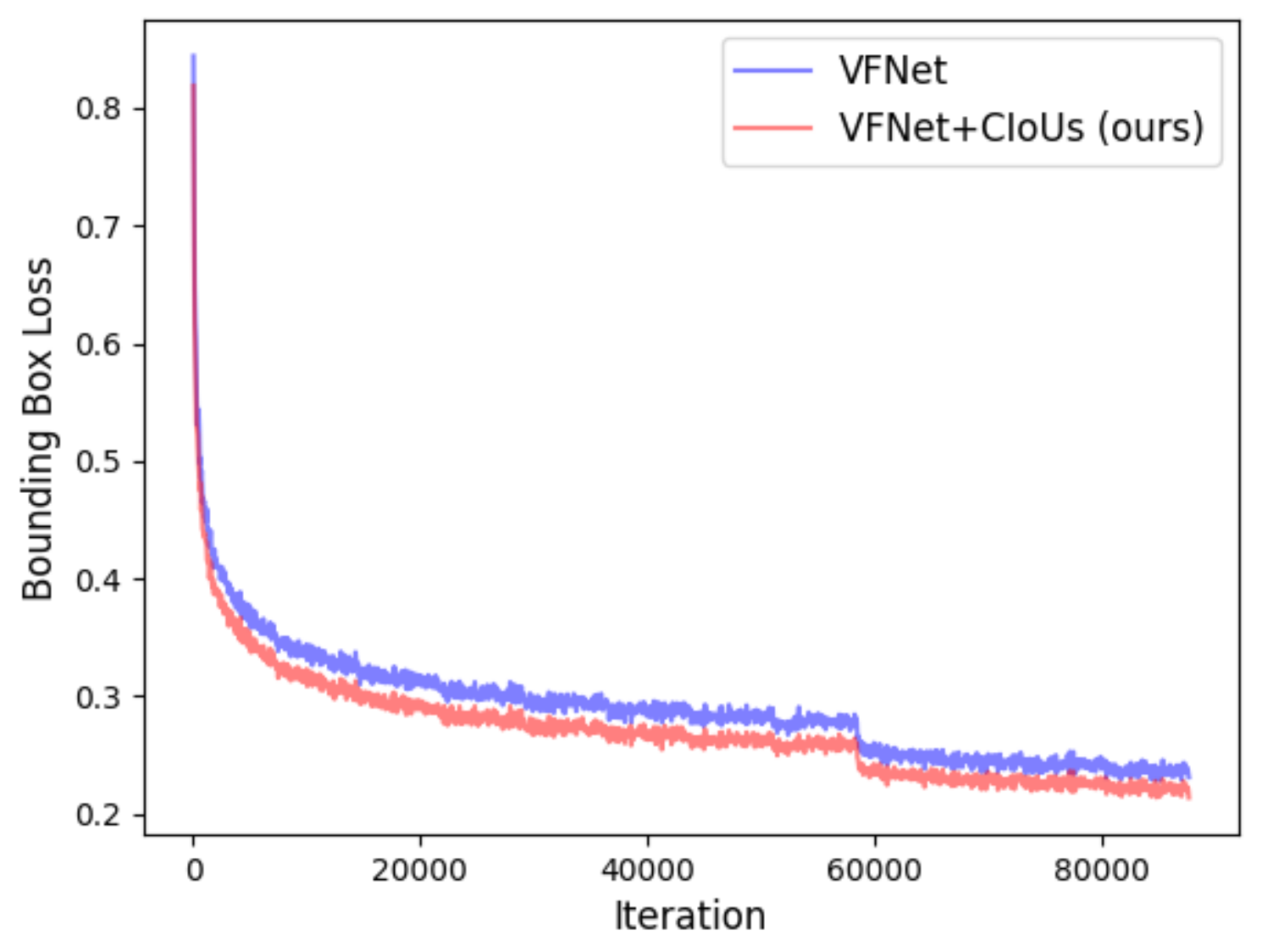}
\caption{The bounding box loss of VFNet and VFNet+CIoUs.}
\label{fig:4}
\end{figure}

From Table \ref{table:4}, we can see that after applying our proposed method to the GFLV2 and VFNet, the two state-of-the-art models are further improved on COCO \textit{val2017}. The improvement comes from the dynamic label assignment strategy (ATSS with our method) which selects samples with high quality predicted boxes as the positives. Thus, the network could focus on those positives and further refine them. Figures \ref{fig:3} and \ref{fig:4} illustrate and compare the bounding box losses of GFLV2 to GFLV2+CIoUs, and VFNet to VFNet+CIoUs, respectively. Similar to the application of our method to ATSS, the models with our method yield lower bounding box losses with the training proceeding, which demonstrates that our proposed method could help the detection models to select higher-quality predicted bounding boxes that have lower bounding box losses.

\begin{table}[H]
\centering
\caption{The application of the proposed approach to state-of-the-art models.}
\newcolumntype{C}{>{\centering\arraybackslash}X}
\begin{tabularx}{\textwidth}{CCCCCCC}
\toprule
\textbf{Model} & \textbf{AP} & \textbf{AP50} & \textbf{AP75} & \textbf{APs} & \textbf{APm} & \textbf{APl} \\
\midrule
GFLV2 & 40.6 & 58.1 & 44.4 & 22.9 & 44.2 & 52.6 \\
\midrule
GFLV2+CIoUs & \textbf{41.1 
} & 58.6 & 44.8 & 23.7 & 44.3 & 53.9 \\
\midrule
VFL & 41.3 & 59.2 & 44.8 & 24.5 & 44.9 & 54.2 \\
\midrule
VFL+CIoUs & \textbf{41.6} & 59.5 & 44.9 & 24.3 & 45.1 & 54.6 \\
\bottomrule

\end{tabularx}
\label{table:4}
\end{table}
\vspace{-12pt}
\vspace{-12pt}




\subsection{Comparison to the State-of-the-Art}


We apply our dynamic label assignment strategy (CIoUs+QFL+IOU branch) to the COCO \textit{test-dev} \cite{lin2014microsoft} and compare them with the original ATSS. Table \ref{table:5} illustrates the experimental results of the state-of-the-art models and our methods, where ``MStrain'' indicates multi-scale training strategy and ``*'' denotes our re-implementation. Scheduler 1x indicates 90 k iterations (12 epochs) and scheduler 2x represents 180 k iterations (24~epochs) on COCO \textit{test-dev} benchmark. 

\startlandscape
\tablesize{\footnotesize}
\begin{table}[H]
\caption{Evaluation results on COCO test-dev.}
\newcolumntype{C}{>{\centering\arraybackslash}X}
\begin{tabularx}{\textwidth}{CCCCCCCCCCC}
\toprule
\textbf{Method} & \textbf{Backbone} & \textbf{Scheduler} & \textbf{Time} & \textbf{MStrain} & \textbf{AP} & \textbf{AP50} & \textbf{AP75} & \textbf{APs} & \textbf{APm} & \textbf{APl} \\
\midrule
FreeAnchor \cite{zhang2019freeanchor} & ResNet-101 & 2x & - & \checkmark & 43.1 & 62.2 & 46.4 & 24.5 & 46.1 & 54.8 \\

FreeAnchor \cite{zhang2019freeanchor} & ResNeXt-101-32x8d & 2x & - & \checkmark & 44.9 & 64.3 & 48.5 & 26.8 & 48.3 & 55.9 \\

TridentNet \cite{li2019scale} & ResNet-101 & 2x & - & \checkmark & 42.7 & 63.6 & 46.5 & 23.9 & 46.6 & 56.6 \\

FCOS \cite{tian2019fcos} & ResNet-101 & 2x & - & \checkmark & 41.5 & 60.7 & 45.0 & 24.4 & 44.8 & 51.6 \\

FCOS \cite{tian2019fcos} & ResNeXt-101-64x4d & 2x & - & \checkmark & 44.7 & 64.1 & 48.4 & 27.6 & 47.5 & 55.6 \\

SAPD \cite{zhu2020soft} & ResNet-101 & 2x & - & \checkmark & 43.5 & 63.6 & 46.5 & 24.9 & 46.8 & 54.6 \\

SAPD \cite{zhu2020soft} & ResNet-101-DCN & 2x & - & \checkmark & 46.0 & 65.9 & 49.6 & 26.3 & 49.2 & 59.6 \\

RepPoints \cite{yang2019reppoints} & ResNet-101 & 2x & - & & 41.0 & 62.9 & 44.3 & 23.6 & 44.1 & 51.7 \\

RepPoints \cite{yang2019reppoints} & ResNet-101-DCN & 2x & - & \checkmark & 45.0 & 66.1 & 49.0 & 26.6 & 48.6 & 57.5 \\
\midrule

GFL \cite{li2020generalized} & ResNet-101 & 2x & - & \checkmark & 45.0 & 63.7 & 48.9 & 27.2 & 48.8 & 54.5 \\

GFL \cite{li2020generalized} & ResNet-101-DCN & 2x & - & \checkmark & 47.3 & 66.3 & 51.4 & 28.0 & 51.1 & 59.2 \\
\midrule

ATSS * 
\cite{zhang2020bridging} & ResNet-50 & 1x & 80 ms & & 39.2 & 57.5 & 42.6 & 22.3 & 41.9 & 49.0 \\

ATSS * \cite{zhang2020bridging} & ResNet-50-DCN & 1x & 96 ms & & 43.0 & 61.2 & 46.8 & 24.5 & 45.9 & 55.3 \\

ATSS \cite{zhang2020bridging} & ResNet-101 & 2x & 105 ms & \checkmark & 43.6 & 62.1 & 47.4 & 26.1 & 47.0 & 53.6 \\

ATSS \cite{zhang2020bridging} & ResNet-101-DCN & 2x & 131 ms & \checkmark & 46.3 & 64.7 & 50.4 & 27.7 & 49.8 & 58.4 \\


ATSS \cite{zhang2020bridging} & ResNeXt-64x4d-101 & 2x & 191 ms & \checkmark & 45.6 & 64.6 & 49.7 & 28.5 & 48.9 & 55.6 \\

ATSS \cite{zhang2020bridging} & ResNeXt-32x8d-101-DCN & 2x & 225 ms & \checkmark & 47.7 & 66.6 & 52.1 & 29.3 & 50.8 & 59.7 \\

ATSS \cite{zhang2020bridging} & ResNeXt-64x4d-101-DCN & 2x & 236 ms & \checkmark & 47.7 & 66.5 & 51.9 & 29.7 & 50.8 & 59.4 \\


\midrule
\textbf{Dynamic ATSS 
} & ResNet-50 & 1x & 80 ms & & 40.3 & 57.9 & 44.1 & 22.5 & 43.6 & 51.2 \\

\textbf{Dynamic ATSS} & ResNet-50-DCN & 1x & 96 ms & & 44.4 & 61.9 & 48.6 & 25.1 & 47.8 & 58.1 \\

\textbf{Dynamic ATSS} & ResNet-101 & 2x & 105 ms & \checkmark & 44.7 & 62.5 & 48.9 & 26.7 & 48.3 & 55.7 \\

\textbf{Dynamic ATSS} & ResNet-101-DCN & 2x & 131 ms & \checkmark & 47.3 & 65.0 & 51.7 & 28.3 & 50.8 & 60.4 \\


\textbf{Dynamic ATSS} & ResNeXt-64x4d-101 & 2x & 191 ms & \checkmark & 46.5 & 64.7 & 50.8 & 29.1 & 49.7 & 57.6 \\

\textbf{Dynamic ATSS} & ResNeXt-32x8d-101-DCN & 2x & 225 ms & \checkmark & 48.6 & 66.7 & 53.0 & 29.9 & 51.5 & 61.8 \\

\textbf{Dynamic ATSS} & ResNeXt-64x4d-101-DCN & 2x & 236 ms & \checkmark & 48.6 & 66.7 & 52.9 & 29.4 & 51.9 & 61.3 \\
\bottomrule

\end{tabularx}
\label{table:5}
\end{table}
\finishlandscape

Table \ref{table:5} illustrates the experimental results of some state-of-the-art models and our methods. The NVIDIA Tesla P100 GPU is employed to test the inference time. From  Table \ref{table:5}, we can see that the proposed method does not introduce extra cost and the real processing time for each image per GPU is the same as the original model (e.g., 80 ms indicates processing one image per GPU needs 80 milliseconds). 
Our Dynamic ATSS model could boost the original ATSS by around 1\% in overall performance and 2\% for large objects (APl). We also notice that our results are similar to state-of-the-art GFL \cite{li2020generalized}, which is also based on ATSS for label assignment. While due to the introduction of distribution focal loss (DFL), GFL \cite{li2020generalized} predicts the bounding boxes multiple times for each sample, which would slightly increase the number of parameters. Our model is much simpler without introducing extra computational cost and parameters, and directly predicts the 4 offsets of the bounding boxes, just as the original ATSS.

\section{Conclusions}
In this paper, we have illustrated the advantages of using predictions for defining positive samples and negative samples for object detection models and proposed a simple and effective method to improve the performance of adaptive label assignment algorithms. By combining the predicted IoUs and anchor IoUs, the label assignment approaches could divide the positive and negative samples dynamically according to the predictions. The predicted IoUs could be combined with the anchor IoUs (prior) by summation, so the dynamic ATSS could implement label assignment based on training status and select the samples with higher quality predicted bounding boxes as the positives. When appling the proposed approach to state-of-the-art detection models with the adaptive algorithm for label assignment, the performance of the models could be further improved.

The proposed method can select samples with high quality predicted bounding boxes as the positive samples even though they have relatively low quality anchors (with low IoUs between the anchors and the ground truth bounding boxes). While most detection models divide the positive samples and negative samples based on fixed anchor boxes, those anchor boxes are always positive samples during the training process only when their IoUs to the ground truth boxes are relatively high, even though the predicted boxes are not that excellent. Our method introduces the training status (the predicted box for each sample) into the division of positives and negatives, those candidates that easily generate high quality predicted boxes are easier to be categorized as the positives, which would assist the network to focus on those high quality samples and generate more high quality bounding boxes. In addition, the proposed method does not introduce extra computational costs and the inference time remains the same as the original methods, while our model increases the accuracy of the detection.

Since our method forces the network to pay much attention to the samples with high quality predicted boxes, the improvement for AP50 and small objects is less than that for AP75 and large objects. We are working to improve the algorithm to further boost its performance for AP50 and small objects. Prediction-based label assignment algorithms for object detection are significant to select high-quality positive samples according to training status. We hope our simple and effective approach could inspire more work on designing dynamic object detectors based on training status.

\vspace{6pt}



\authorcontributions{
T.Z.: Conceptualization, Methodology, Software, Writing, Editing. B.L.: Supervision, Writing, Review.  A.S.: Supervision, Writing, Review. G.W.: Conceptualization, Supervision, Writing, Review, Editing. All authors have read and agreed to the published version of the manuscript. 
}

\funding{
This work was partly supported in part by the Natural Sciences and Engineering Research Council of Canada (NSERC) under grant no. RGPIN-2021-04244, the United States Department of Agriculture (USDA) under grant no. 2019-67021-28996. 
}

\institutionalreview{
Not applicable. 
}

\informedconsent{
Not applicable. 
}

\dataavailability{
The source code is available at 
\url{https://github.com/ZTX-100/DLA-Combined-IoUs}. 
}


\conflictsofinterest{
The authors declare no conflict of interest. 
}


\begin{adjustwidth}{-\extralength}{0cm}

\reftitle{References}

\end{adjustwidth}
\end{document}